\documentclass[10pt,twocolumn,letterpaper]{article}

\usepackage[pagenumbers]{cvpr} 
\usepackage{pifont}
\usepackage{marvosym}
\usepackage{booktabs}
\usepackage{multirow}
\usepackage{makecell}
\usepackage{adjustbox}
\usepackage{colortbl}
\usepackage{tabularx}

\definecolor{cvprblue}{rgb}{0.21,0.49,0.74}
\usepackage[pagebackref,breaklinks,colorlinks,allcolors=cvprblue]{hyperref}

\def\paperID{*****} 
\def\confName{CVPR}
\def\confYear{2026}

\title{LOGER: Local--Global Ensemble for Robust Deepfake Detection in the Wild}

\author{ 
\small Fei Wu$^{1}$, Dagong Lu$^{2}$, Mufeng Yao$^{2}$, Xinlei Xu$^{2}$,  Fengjun Guo$^{2,\text{\Letter}}$ 
\\
$^1$ \small Shanghai Jiao Tong University \\
$^2$ \small INTSIG Information \\
}

\begin{document}
\maketitle
\let\thefootnote\relax\footnotetext{$^{\text{\Letter}}$ Corresponding author.}
\begin{abstract}

Robust deepfake detection in the wild remains challenging due to the ever-growing variety of manipulation techniques and uncontrolled real-world degradations.
Forensic cues for deepfake detection reside at two complementary levels: global-level anomalies in semantics and statistics that require holistic image understanding, and local-level forgery traces concentrated in manipulated regions that are easily diluted by global averaging.
Since no single backbone or input scale can effectively cover both levels, we propose LOGER, a \textbf{LO}cal--\textbf{G}lobal \textbf{E}nsemble framework for \textbf{R}obust deepfake detection.
The global branch employs heterogeneous vision foundation model backbones at multiple resolutions to capture holistic anomalies with diverse visual priors.
The local branch performs patch-level modeling with a Multiple Instance Learning top-$k$ aggregation strategy that selectively pools only the most suspicious regions, mitigating evidence dilution caused by the dominance of normal patches; dual-level supervision at both the aggregated image level and individual patch level keeps local responses discriminative.
Because the two branches differ in both granularity and backbone, their errors are largely decorrelated, a property that logit-space fusion exploits for more robust prediction.
LOGER achieves 2nd place in the NTIRE 2026 Robust Deepfake Detection Challenge, and further evaluation on multiple public benchmarks confirms its strong robustness and generalization across diverse manipulation methods and real-world degradation conditions.

\end{abstract}

\section{Introduction}
\label{sec:intro}

Modern face manipulation methods produce deepfakes realistic enough to fool both humans and automated systems~\cite{Goodfellow2014GAN,Rombach_2022_CVPR}, fueling growing concerns about identity fraud, political disinformation, and financial scams~\cite{Chesney2019Deepfakes,Tolosana2020Survey}.
While detecting a known manipulation type under controlled conditions is largely solved, the real challenge is generalization: deployed detectors must handle unseen generators, diverse manipulation categories, and unpredictable post-processing that content undergoes in the wild.
Deepfakes are rarely encountered in pristine form: media is routinely compressed, resized, or blurred during transmission, and adversaries may intentionally degrade quality to suppress forgery artifacts.
The NTIRE 2026 Robust Deepfake Detection Challenge~\cite{ntire26deepfake} targets exactly this setting, requiring detectors to remain accurate under diverse manipulations and realistic degradations.

Early detectors relied on low-level artifacts such as blending boundaries~\cite{li2020face}, texture irregularities~\cite{zhao2021multi}, and frequency-domain anomalies~\cite{Durall2020Frequency,Frank2020FrequencyBias}, but these cues degrade quickly under JPEG compression and spatial resizing~\cite{tan2024rethinking}.
Subsequent work on data alignment~\cite{chen2025dualdataalignment,guillaro2025bias}, latent-space augmentation~\cite{yan2024transcending}, and orthogonal decomposition~\cite{yan2024effort} improves cross-generator transfer, yet scaling-law studies show that detection error decreases only as a power law of data diversity, meaning improvements slow down rapidly and a single model trained on limited sources faces inherent generalization ceilings~\cite{wang2025scaling}.
Vision foundation models (VFMs) offer a promising alternative.
Pre-trained on web-scale data with self-supervised or contrastive objectives, VFMs encode both low-level texture patterns and high-level semantic structures that are highly discriminative for distinguishing real from generated content.
Models such as DINOv3~\cite{simeoni2025dinov3} and MetaCLIP2~\cite{chuang2025meta}, even with minimal adaptation, outperform purpose-built detectors on in-the-wild benchmarks~\cite{yermakov2026deepfake,zhou2025brought}, confirming that these pre-trained representations transfer well to the forensic domain.

However, VFM-based full-image classification still has a fundamental limitation: it averages features over the entire image, capturing holistic anomalies (e.g., unnatural facial symmetry) but diluting the subtle forgery traces that concentrate in small manipulated regions~\cite{li2020sharp}.
Conversely, patch-level analysis can detect such fine-grained local artifacts but lacks the global context needed to recognize semantic-level inconsistencies.
Since forensic cues reside at both global and local levels, combining detection strategies at different granularities with different backbones is more promising than perfecting any single strategy.

We propose \textbf{LOGER}, a two-branch local--global ensemble framework that acts on this observation.
The global branch leverages heterogeneous VFM backbones at multiple resolutions for holistic anomaly detection, while the local branch uses MIL top-$k$ aggregation to focus on the most suspicious patches with dual-level supervision.
Because the two branches differ in both granularity and backbone, their errors are largely decorrelated, which logit-space fusion exploits for more stable predictions.

Our main contributions are as follows:
\begin{itemize}
	\item We propose LOGER, a two-branch local--global ensemble framework whose global branch leverages heterogeneous VFM backbones at multiple resolutions for holistic anomaly detection while the local branch focuses on suspicious regions, providing comprehensive forensic coverage of both global and local forgery traces for more accurate detection.
	\item We design a local detection branch with MIL top-$k$ aggregation that mitigates evidence dilution by selectively pooling only the most suspicious patches, jointly supervised at both the aggregated and individual patch levels to ensure accurate image-level predictions while maintaining discriminative local responses.
	\item We adopt logit-space fusion that preserves each model's full confidence range. LOGER achieves 2nd place in the NTIRE 2026 Robust Deepfake Detection Challenge~\cite{ntire26deepfake} and generalizes well across multiple public benchmarks.
\end{itemize}

\section{Related Work}
\label{sec:related}

\subsection{Deepfake Detection Datasets}

Benchmark datasets have shaped deepfake detection research by defining increasingly challenging evaluation settings.
FaceForensics++~\cite{rossler2019faceforensics} established the first large-scale benchmark with multiple manipulation techniques and compression levels, and Celeb-DF~\cite{li2020celeb} subsequently exposed severe cross-dataset performance drops, making generalization a central concern.

Later benchmarks extend evaluation along two axes.
The first is \emph{scale and diversity}: DFDC~\cite{Dolhansky2020DFDC} provides large-scale real-world diversity, DF40~\cite{yan2024df40} covers 40 manipulation methods, HydraFake~\cite{tan2025veritas} defines a hierarchical out-of-distribution evaluation protocol, and ScaleDF~\cite{wang2025scaling} reveals power-law scaling between data diversity and detection accuracy.
The second is \emph{robustness}: DeeperForensics~\cite{jiang2020deeperforensics} introduces controlled perturbations, WildDeepfake~\cite{zi2020wilddeepfake} captures noise from network propagation, and MFFI~\cite{miao2025mffi} targets multi-level degradation.
A consistent lesson is that data diversity, rather than sheer quantity, is the primary driver of generalization~\cite{wang2025scaling}, a principle that informs our multi-source training strategy.

\subsection{Deepfake Detection Methods}

Deepfake detection methods can be broadly grouped into three categories based on the type of representation they exploit.

\paragraph{Artifact-based methods.}
Early approaches target specific forgery artifacts: blending boundaries~\cite{li2020face}, texture irregularities~\cite{zhao2021multi}, and frequency-domain anomalies~\cite{Durall2020Frequency,Frank2020FrequencyBias}.
Data-synthesis strategies such as SBI~\cite{shiohara2022detecting} further improve generalization by creating pseudo-fakes during training.
While effective on controlled benchmarks, these hand-crafted cues are sensitive to post-processing and do not transfer well across generators.

\paragraph{Generalization-oriented methods.}
A second category focuses on closing the domain gap between training and unseen generators.
Data alignment (DDA~\cite{chen2025dualdataalignment}, B-Free~\cite{guillaro2025bias}) and latent-space augmentation (LSDA~\cite{yan2024transcending}) reduce distribution-specific biases.
Orthogonal decomposition (Effort~\cite{yan2024effort}) decomposes the feature space into orthogonal subspaces, preserving pre-trained knowledge while learning forgery-specific cues.
These methods improve cross-generator transfer but still face inherent generalization ceilings when training data diversity is limited~\cite{wang2025scaling}.

\paragraph{Foundation-model-based methods.}
The most recent line of work adapts VFMs to forensics.
Forensics Adapter~\cite{cui2025forensics} and LoRA-based methods~\cite{kong2023enhancing,kong2024moe} fine-tune pre-trained CLIP representations with minimal extra parameters.
Multi-source training pipelines (D$^3$~\cite{yang2025d}, GenD~\cite{yermakov2026deepfake}) pair diverse generators with parameter-efficient adaptation; notably, GenD demonstrates that tuning only the Layer Normalization parameters of a VFM backbone can achieve competitive cross-benchmark performance.
MLLM-based detectors such as Veritas~\cite{tan2025veritas}, FakeShield~\cite{xu2024fakeshield}, and Legion~\cite{kang2025legion} add semantic reasoning for more explainable predictions, though their robustness on degraded inputs still lags behind specialized visual approaches.

Despite this progress, cross-benchmark evaluations consistently show that no single detection strategy is sufficient~\cite{wang2025scaling,DeepfakeBench}.
Our work combines local and global branches with heterogeneous VFM backbones, exploiting their complementary strengths through logit-space fusion.

\begin{figure*}[!t]
    \centering
    \includegraphics[width=\linewidth]{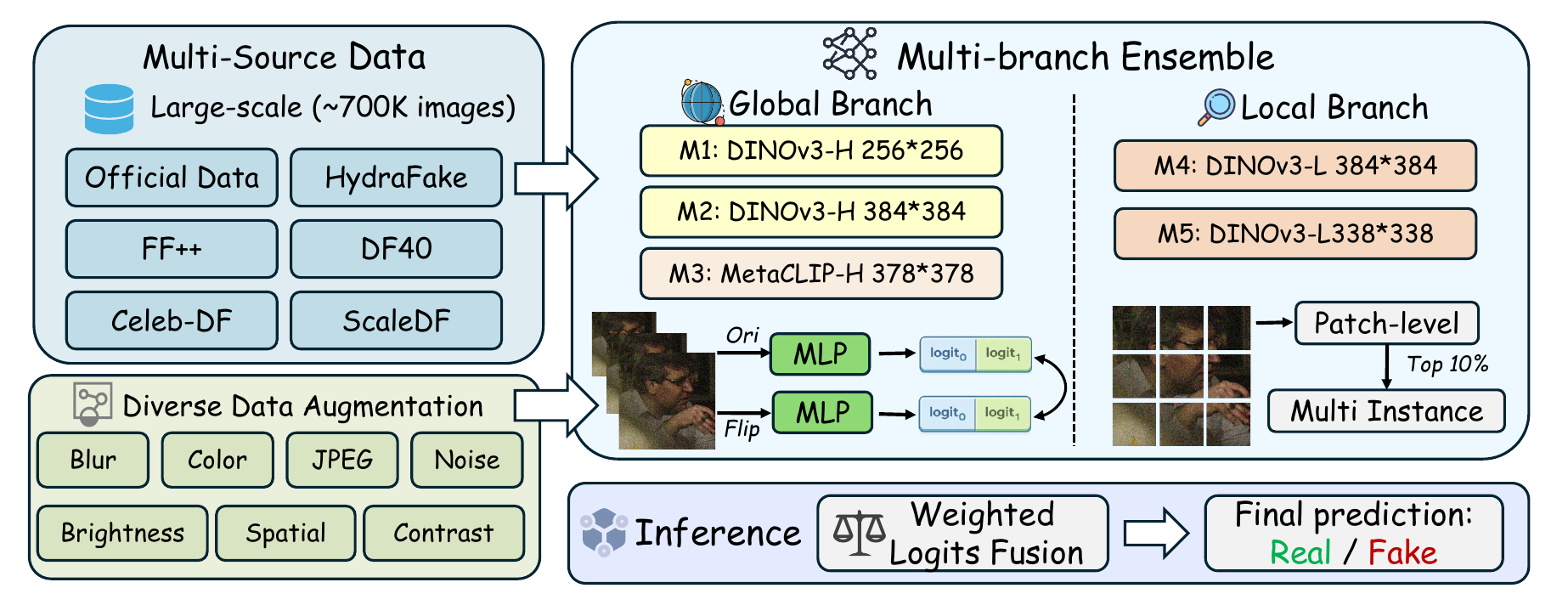}
    \caption{Overview of the proposed LOGER framework. Training data are sampled from a multi-source candidate pool with diverse degradation augmentation. The global branch performs full-image detection using DINOv3-H (M1, M2) and MetaCLIP2-H (M3) at multiple resolutions. The local branch employs DINOv3-L (M4, M5) with patch-level modeling and top-10\% MIL pooling. All outputs are fused via logit-space averaging with test-time augmentation.}
    \label{fig:overview}
\end{figure*}

\subsection{Impact of Degradation}

Deployed deepfake content inevitably undergoes JPEG compression, blur, and resolution reduction, disrupting the fine-grained cues that detectors rely on~\cite{rossler2019faceforensics,jiang2020deeperforensics}.
Strong compression suppresses high-frequency forensic traces~\cite{Durall2020Frequency}, and resolution loss distorts local textures critical for artifact-based methods~\cite{Tolosana2020Survey}.
Recent studies show that even state-of-the-art detectors experience significant performance drops when evaluated on compressed or resized inputs compared to pristine conditions~\cite{tan2024rethinking,yan2024df40}.
Degradation-aware augmentation~\cite{miao2025mffi} and robustness benchmarking~\cite{jiang2020deeperforensics} partially address the gap, yet scaling-law analyses show that augmentation alone cannot fully compensate and remains essential even at large data scales~\cite{wang2025scaling}.
These findings motivate our multi-scale input processing, heterogeneous backbones, and aggressive degradation augmentation, which together improve resilience across a wide range of real-world post-processing conditions.

\section{Method}
\label{sec:method}

\subsection{Overview}
\label{sec:overview}

LOGER adopts a two-branch design to jointly capture global-level and local-level forensic cues (Fig.~\ref{fig:overview}).
The \textbf{global branch} performs full-image detection using three models with heterogeneous backbones or inference resolutions (M1--M3),
while the \textbf{local branch} performs patch-level forensic analysis using two models with MIL-based aggregation (M4--M5).
All outputs are fused in the logit space with optional test-time augmentation.
To maximize generator diversity, we build a multi-source candidate pool (approximately 700K images) from several public deepfake benchmarks and sample a subset for each model's training (see Sec.~\ref{sec:setup} for per-model configurations).
All models further apply degradation augmentation that randomly composes perturbations from multiple groups to simulate real-world post-processing conditions (details in Sec.~\ref{sec:setup}).
We describe each component below.

\subsection{Global Branch}
\label{sec:global}

The global branch performs full-image detection at multiple scales with diverse backbones to capture both semantic inconsistencies and statistical artifacts.

\paragraph{Multi-scale DINOv3 detection.}
M1 and M2 share the same DINOv3-Huge backbone~\cite{simeoni2025dinov3} with full-parameter fine-tuning.
We use full-parameter fine-tuning rather than parameter-efficient alternatives (e.g., LoRA~\cite{kong2023enhancing} or linear probing), as the gap between the VFM pre-training objective and forensic binary classification is large enough that partial adaptation tends to underfit forgery-relevant cues.
M1 is trained and inferred at the same resolution, while M2 is trained at the same resolution as M1 but inferred at a higher resolution, so that fine-grained forensic details (e.g., subtle boundary artifacts or compression traces) that a single fixed resolution might discard are preserved at one scale or the other.
Each backbone is followed by a two-layer MLP classification head ($d_{\text{in}}\!\rightarrow\!256\!\rightarrow\!2$) with ReLU activation and Dropout(0.1).
Both models are trained with Focal Loss~\cite{lin2017focal}, which down-weights well-classified examples and focuses gradient on hard samples, improving discrimination on ambiguous or degraded inputs.

\paragraph{Backbone heterogeneity via MetaCLIP2.}
M3 uses MetaCLIP2-Huge~\cite{chuang2025meta} as the backbone.
DINOv3 is trained via self-supervised objectives on image patches, encoding rich spatial and structural representations that are sensitive to local texture and boundary irregularities.
MetaCLIP2, in contrast, is trained through contrastive image-text alignment, learning visual features grounded in high-level semantic concepts such as object identity and scene context.
Because their pre-training objectives differ fundamentally, the two backbones encode complementary priors: DINOv3 is more sensitive to spatial and structural details while MetaCLIP2 brings stronger high-level semantic grounding, and their combination reduces correlated errors in the ensemble.
M3 uses a staged loss schedule: cross-entropy for the first 20\% of training to ensure stable convergence, then Focal Loss for the remaining 80\% to shift focus toward hard samples.

\subsection{Local Branch}
\label{sec:local}

Face swapping and local reenactment often leave forgery traces concentrated in specific regions (e.g., blending boundaries, eye/mouth areas).
Global detectors average over all spatial locations and therefore suffer from evidence dilution when strong signals from a few forged patches are overwhelmed by numerous normal ones~\cite{li2020sharp}.
The local branch counters this by operating at the patch level and aggregating only the most suspicious regions.

\paragraph{Patch-level modeling and MIL top-$k$ aggregation.}
M4 and M5 use DINOv3-Large~\cite{simeoni2025dinov3} as the backbone, since patch-level modeling produces substantially more tokens than full-image classification and the Large variant offers a better efficiency trade-off.
The input image is split into a grid of non-overlapping patches, and each patch feature is mapped to real/fake logits via lightweight classification layers, yielding a response map $\{d_i\}_{i=1}^{N}$ where $N$ is the patch count and $d_i = l_i^{\text{fake}} - l_i^{\text{real}}$ is the logit difference of patch $i$.
We aggregate patch-level predictions into an image-level score using Multiple Instance Learning (MIL) top-$k$ pooling.
Only the top $k = \lfloor 0.1 \cdot N \rfloor$ patches with the highest fake scores are selected for pooling:
\begin{equation}
\label{eq:mil}
d_{\text{img}} = \frac{1}{k} \sum_{i \in \mathcal{S}_k} d_i, \quad \mathcal{S}_k = \operatorname{top\text{-}k}(\{d_i\}_{i=1}^{N}),
\end{equation}
where $\mathcal{S}_k$ denotes the index set of the $k$ highest-scoring patches.
By focusing on the most suspicious regions, this strategy enhances sensitivity to small forged areas while suppressing noise from normal patches.

\paragraph{Training objective and multi-resolution inference.}
M4 and M5 are optimized with a weighted multi-term objective:
\begin{equation}
\label{eq:local_loss}
\mathcal{L}_{\text{local}} = \mathcal{L}_{\text{CE}} + 0.5\,\mathcal{L}_{\text{AUC}} + 0.5\,\mathcal{L}_{\text{MIL}} + \mathcal{L}_{\text{reg}},
\end{equation}
where $\mathcal{L}_{\text{CE}}$ and $\mathcal{L}_{\text{AUC}}$ are both computed on the aggregated image-level score $d_{\text{img}}$ from Eq.~\ref{eq:mil}: $\mathcal{L}_{\text{CE}}$ is the binary cross-entropy loss applied to $\sigma(d_{\text{img}})$, and $\mathcal{L}_{\text{AUC}}$ is a pairwise AUC surrogate loss that maximizes the margin between positive and negative samples to directly optimize ranking quality.
$\mathcal{L}_{\text{MIL}}$ is a patch-level MIL loss applied on the selected top-$k$ patch scores $\{d_i\}_{i\in\mathcal{S}_k}$, encouraging each individual patch to produce discriminative local responses rather than relying solely on the aggregated signal.
$\mathcal{L}_{\text{reg}}$ consists of regularization terms that prevent degenerate solutions where all patch scores collapse to uniformly positive or negative values.
Supervising at both levels keeps individual patch responses discriminative while still producing accurate image-level predictions.
M4 is trained at a lower resolution and inferred at a higher resolution, producing a denser patch grid that allows the model to observe finer spatial details at test time without retraining.
M5 is initialized from M4's best checkpoint and further fine-tuned at a different resolution with reduced learning rates, adapting to a different spatial granularity while retaining the forensic representations learned by M4.
This continuation training strategy produces two complementary local detectors: M4 benefits from the resolution gap between training and inference, while M5 operates at a resolution it was explicitly optimized for.

\subsection{Logit-Space Fusion}
\label{sec:fusion}

We aggregate predictions from all five models in the logit space rather than averaging output probabilities.
For each global-branch model $m$, we convert its two-class logits into a single directional evidence score $d_m = l_m^{\text{fake}} - l_m^{\text{real}}$; for each local-branch model, $d_m$ is directly set to the aggregated score $d_{\text{img}}$ from Eq.~\ref{eq:mil}, which is already a logit difference.
This formulation retains only the directional evidence for or against the fake class, providing a comparable scale across models regardless of individual logit magnitudes.
The fused evidence and final fake probability are computed via averaging:
\begin{equation}
\label{eq:fusion}
\bar{d}(x) = \sum_{m=1}^{M} \alpha_m\, d_m(x), \quad p(x) = \sigma\!\left(\bar{d}(x)\right),
\end{equation}
where $M{=}5$, $\sigma$ is the sigmoid function, and we use uniform weights $\alpha_m = 1/M$.
We prefer logit-space fusion over probability averaging because the sigmoid compresses large logit differences into near-saturated probabilities, masking inter-model disagreements.
Fusing before the sigmoid retains each model's full confidence range, allowing a confident minority opinion to influence the final decision.

\begin{table*}[!t]
\centering
\caption{Training and inference configurations for all five models. LR$_b$ and LR$_h$ denote the learning rates for the backbone and classification head, respectively.}
\vspace{-2mm}
\begin{adjustbox}{max width=\textwidth}
\begin{tabular}{ccccccccccccc}
\toprule
Model & Branch & Backbone & Fine-tune & Train Res. & Infer Res. & LR$_b$ & LR$_h$ & Loss & Epochs & TTA & GPUs \\
\midrule
M1 & Global & DINOv3-H & Full & $256\!\times\!256$ & $256\!\times\!256$ & 8e-5 & 2e-3 & Focal & 10 & Flip & 8$\times$H800 \\
M2 & Global & DINOv3-H & Full & $256\!\times\!256$ & $384\!\times\!384$ & 8e-5 & 2e-3 & Focal & 10 & Flip & 8$\times$H800 \\
M3 & Global & MetaCLIP2-H & Full & $378\!\times\!378$ & $378\!\times\!378$ & 1e-4 & 1e-4 & CE$\to$Focal & 20 & -- & 16$\times$A10 \\
M4 & Local & DINOv3-L & Full & $224\!\times\!224$ & $384\!\times\!384$ & 1e-4 & 5e-6 & CE+AUC+MIL & 20 & Flip & 16$\times$A10 \\
M5 & Local & DINOv3-L & From M4 & $338\!\times\!338$ & $338\!\times\!338$ & 1e-5 & 5e-7 & CE+AUC+MIL & 20 & Flip & 16$\times$A10 \\
\bottomrule
\end{tabular}
\end{adjustbox}
\label{tab:training_config}
\vspace{-2mm}
\end{table*}

\paragraph{Test-time augmentation.}
For M1, M2, M4, and M5, we apply horizontal-flip TTA: each model processes both the original image and its horizontally flipped version, and their logits are averaged before entering the fusion.
Since horizontal flipping does not change a face's authenticity, the detector should yield consistent predictions for both views; averaging the two reduces variance from orientation-dependent artifacts.

\section{Experiments}
\label{sec:experiments}

\subsection{Experimental Setup}
\label{sec:setup}

\paragraph{Datasets.}
The official competition training set~\cite{ntire26deepfake} contains only 1,000 images, which is insufficient to cover the wide variety of manipulation types and degradation conditions encountered at test time.
To provide broader generator diversity, we construct a multi-source candidate pool (approximately 700K images) from several public deepfake benchmarks, including HydraFake~\cite{tan2025veritas}, FaceForensics++~\cite{rossler2019faceforensics}, DF40~\cite{yan2024df40}, Celeb-DF~\cite{li2020celeb}, and ScaleDF~\cite{wang2025scaling}.
For M1, M2, M4, and M5, the training data consist of 1,000 official samples plus 20K images randomly sampled from this pool.
For M3, we sample 150K images from the official set, DF40, and HydraFake to expose the MetaCLIP2 backbone to a broader range of forgery types and increase the diversity of training sources within the ensemble.
All models apply competition-provided degradation augmentation~\cite{codabench12761}, which randomly composes perturbations from multiple groups (blur, noise, JPEG compression, color shift, spatial distortion) to simulate real-world post-processing.

\paragraph{Implementation details.}
All models are implemented in PyTorch 2.3.1 with timm 1.0.20.
The detailed per-model training configurations are summarized in Table~\ref{tab:training_config}.
All models use AdamW ($\beta_1{=}0.9$, $\beta_2{=}0.999$, weight decay $10^{-2}$) with discriminative learning rates for the backbone and classification head, and gradient clipping at max norm 1.0.
All models adopt a WeightedRandomSampler to address class imbalance in the training set.
For the global branch, M1 and M2 share the same checkpoint trained at 256$\times$256, with M1 inferred at 256$\times$256 and M2 inferred at 384$\times$384; M3 is trained and inferred at 378$\times$378.
For the local branch, M4 is trained at 224$\times$224 and inferred at 384$\times$384; M5 is initialized from M4 and fine-tuned at 338$\times$338 with reduced learning rates.
On a single H800 GPU, the five-model ensemble processes images at approximately 17 FPS in steady state, making it practical for offline forensic analysis.

\paragraph{Evaluation metrics and comparative methods.}
We report AUC as the primary evaluation metric.
For cross-dataset evaluation, we cite video-level AUC results reported in Effort~\cite{yan2024effort} and GenD~\cite{yermakov2026deepfake} as baselines.
On the NTIRE 2026 competition validation set, we report image-level AUC and compare against representative methods:
Effort~\cite{yan2024effort}, DDA~\cite{chen2025dualdataalignment}, MIRROR~\cite{liu2026mirror}, and GenD~\cite{yermakov2026deepfake} are evaluated using their officially released pre-trained weights;
CLIP-ViT-L~\cite{clip}, ConvNeXt-Large~\cite{liu2022convnet}, and Qwen3-VL~\cite{bai2025qwen3} are trained on the 21K configuration (official + 20K external, same as M1/M2) and evaluated on the validation set;
Reality Defender~\cite{realitydefender} is tested via its public API;
zero-shot commercial MLLMs (GPT-5.2, Claude Sonnet 4.5, Gemini 3 Pro) are queried with a binary real/fake prompt using default parameters.

\begin{table*}[t]
\centering
\caption{Cross-dataset (Protocol-1) and cross-manipulation (Protocol-2) evaluation following Effort~\cite{yan2024effort}. All baseline methods are trained on FF++ c23; LOGER uses multi-source training (Sec.~\ref{sec:setup}). We report video-level AUC (\%). Baseline results are cited from~\cite{yan2024effort}. Best and second-best results are in \textbf{bold} and \underline{underline}. DFo and FFIW are omitted from Protocol-1 as their test data are not easily accessible.}
\vspace{-2mm}
\setlength{\tabcolsep}{3pt}
\begin{adjustbox}{max width=\textwidth}
\begin{tabular}{l|ccccc|c|cccccccc|c}
\toprule
& \multicolumn{6}{c|}{\textbf{Protocol-1: Cross-Dataset}} & \multicolumn{9}{c}{\textbf{Protocol-2: Cross-Manipulation (DF40)}} \\
\cmidrule(lr){2-7} \cmidrule(lr){8-16}
\textbf{Method} & \textbf{CDF-v2} & \textbf{DFD} & \textbf{DFDC} & \textbf{DFDCP} & \textbf{WDF} & \textbf{Avg.} & \textbf{UniFace} & \textbf{BleFace} & \textbf{MobSwap} & \textbf{e4s} & \textbf{FaceDan} & \textbf{FSGAN} & \textbf{InSwap} & \textbf{SimSwap} & \textbf{Avg.} \\
\midrule
F3Net~\cite{qian2020thinking} & 0.789 & 0.844 & 0.718 & 0.749 & 0.728 & 0.766 & 0.809 & 0.808 & 0.867 & 0.494 & 0.717 & 0.845 & 0.757 & 0.674 & 0.746 \\
SPSL~\cite{liu2021spatial} & 0.799 & 0.871 & 0.724 & 0.770 & 0.702 & 0.773 & 0.747 & 0.748 & 0.885 & 0.514 & 0.666 & 0.812 & 0.643 & 0.665 & 0.710 \\
SRM~\cite{luo2021generalizing} & 0.840 & 0.885 & 0.695 & 0.728 & 0.702 & 0.770 & 0.749 & 0.704 & 0.779 & 0.704 & 0.659 & 0.772 & 0.793 & 0.694 & 0.732 \\
CORE~\cite{ni2022core} & 0.809 & 0.882 & 0.721 & 0.720 & 0.724 & 0.771 & 0.871 & 0.843 & 0.959 & 0.679 & 0.774 & \underline{0.958} & 0.855 & 0.724 & 0.833 \\
RECCE~\cite{cao2022end} & 0.823 & 0.891 & 0.696 & 0.734 & 0.756 & 0.780 & 0.898 & 0.832 & 0.925 & 0.683 & 0.848 & 0.949 & 0.848 & 0.768 & 0.844 \\
SLADD~\cite{chen2022self} & 0.837 & 0.904 & 0.772 & 0.756 & 0.690 & 0.792 & 0.878 & 0.882 & 0.954 & 0.765 & 0.825 & 0.943 & 0.879 & 0.794 & 0.865 \\
SBI~\cite{shiohara2022detecting} & 0.886 & 0.827 & 0.717 & 0.848 & 0.703 & 0.796 & 0.724 & 0.891 & 0.952 & 0.750 & 0.594 & 0.803 & 0.712 & 0.701 & 0.766 \\
UCF~\cite{yan2023ucf} & 0.837 & 0.867 & 0.742 & 0.770 & 0.774 & 0.798 & 0.831 & 0.827 & 0.950 & 0.731 & 0.862 & 0.937 & 0.809 & 0.647 & 0.824 \\
IID~\cite{huang2023implicit} & 0.838 & 0.939 & 0.700 & 0.689 & 0.666 & 0.766 & 0.839 & 0.789 & 0.888 & 0.766 & 0.844 & 0.927 & 0.789 & 0.644 & 0.811 \\
LSDA~\cite{yan2024transcending} & 0.875 & 0.881 & 0.701 & 0.812 & 0.797 & 0.813 & 0.872 & 0.875 & 0.930 & 0.694 & 0.721 & 0.939 & 0.855 & 0.793 & 0.835 \\
ProDet~\cite{cheng2024prodet} & 0.926 & 0.901 & 0.707 & 0.828 & 0.781 & 0.829 & 0.908 & \underline{0.929} & \textbf{0.975} & 0.771 & 0.747 & 0.928 & 0.837 & 0.844 & 0.867 \\
CDFA~\cite{lin2024cdfa} & 0.938 & 0.954 & 0.830 & 0.881 & 0.796 & 0.880 & 0.762 & 0.756 & 0.823 & 0.631 & 0.803 & 0.942 & 0.772 & 0.757 & 0.781 \\
Effort~\cite{yan2024effort} & \underline{0.956} & \textbf{0.965} & \underline{0.843} & \underline{0.909} & \underline{0.848} & \underline{0.904} & \underline{0.962} & 0.873 & 0.953 & \underline{0.983} & \underline{0.926} & 0.957 & \textbf{0.936} & \underline{0.926} & \underline{0.940} \\
\midrule
\rowcolor{blue!5}
\textbf{LOGER (Ours)} & \textbf{0.959} & \underline{0.962} & \textbf{0.903} & \textbf{0.943} & \textbf{0.863} & \textbf{0.926} & \textbf{0.991} & \textbf{0.963} & \underline{0.961} & \textbf{0.989} & \textbf{0.950} & \textbf{0.980} & \underline{0.890} & \textbf{0.986} & \textbf{0.964} \\
\bottomrule
\end{tabular}
\end{adjustbox}
\label{tab:cross_effort}
\vspace{-1mm}
\end{table*}

\begin{table}[t]
\centering
\caption{Cross-dataset evaluation following GenD~\cite{yermakov2026deepfake}. All baseline methods are trained on FF++ c23; LOGER uses multi-source training (Sec.~\ref{sec:setup}). We report video-level AUC (\%). Baseline results are cited from~\cite{yermakov2026deepfake}. Best and second-best results are in \textbf{bold} and \underline{underline}.}
\vspace{-2mm}
\fontsize{8pt}{10pt}\selectfont
\setlength{\tabcolsep}{3pt}
\begin{adjustbox}{max width=\linewidth}
\begin{tabular}{l|l|l|ccc}
\toprule
\textbf{Method} & \textbf{Input} & \textbf{Backbone} & \textbf{CDF-v2} & \textbf{DFD} & \textbf{DFDC} \\
\midrule
LipForensics~\cite{haliassos2021lipforensics} & Video & ResNet-18 & 82.4 & -- & 73.5 \\
FTCN~\cite{zheng2021ftcn} & Video & 3D ResNet-50 & 86.9 & -- & 74.0 \\
RealForensics~\cite{haliassos2022realforensics} & Video & Modified CSN & 86.9 & -- & 75.9 \\
SBI~\cite{shiohara2022detecting} & Frame & EfficientNet-B4 & 93.2 & 82.7 & 72.4 \\
AUNet~\cite{zhao2021aunet} & Video & Xception & 92.8 & \textbf{99.2} & 73.8 \\
LSDA~\cite{yan2024transcending} & Frame & EfficientNet-B4 & 91.1 & -- & 77.0 \\
LAA-Net~\cite{nguyen2024laanet} & Frame & EfficientNet-B4 & 95.4 & 98.4 & 86.9 \\
AltFreezing~\cite{wang2023altfreezing} & Video & 3D ResNet-50 & 89.5 & 98.5 & -- \\
NACO~\cite{zhang2024naco} & Video & ViT-B/16 & 89.5 & -- & 76.7 \\
RAE~\cite{tian2024rae} & Frame & ViT-B/16 & 95.5 & \underline{99.0} & 80.2 \\
TALL++~\cite{xu2024tall} & Video & Swin-B & 92.0 & -- & 78.5 \\
ProDet~\cite{cheng2024prodet} & Frame & EfficientNet-B4 & 92.5 & -- & 77.0 \\
UDD~\cite{fu2025udd} & Frame & CLIP ViT-B/16 & 93.1 & 95.5 & 81.2 \\
P\&P~\cite{yan2025plugandplay} & Video & CLIP ViT-L/14 & 94.7 & 96.5 & 84.3 \\
DFD-FCG~\cite{han2025dfdfcg} & Video & CLIP ViT-L/14 & 95.0 & -- & 81.8 \\
ForAda~\cite{cui2025forensics} & Frame & CLIP ViT-L/14 & \underline{95.7} & 97.2 & \underline{87.2} \\
Effort~\cite{yan2024effort} & Frame & CLIP ViT-L/14 & 95.6 & 96.5 & 84.3 \\
GenD~\cite{yermakov2026deepfake} & Frame & DINOv3 ViT-L/16 & 92.2 & 96.6 & 84.7 \\
\midrule
\rowcolor{blue!5}
\textbf{LOGER (Ours)} & Frame & DINOv3+MetaCLIP2 & \textbf{95.9} & 96.2 & \textbf{90.3} \\
\bottomrule
\end{tabular}
\end{adjustbox}
\label{tab:cross_gend}
\vspace{-2mm}
\end{table}

\subsection{Comparison with State-of-the-Art}
\label{sec:sota}

\paragraph{Cross-dataset generalization.}
We follow the evaluation pipelines of Effort~\cite{yan2024effort} and GenD~\cite{yermakov2026deepfake}: faces are detected with RetinaFace, cropped with a 1.3$\times$ margin, and for video-level evaluation, 32 frames are uniformly sampled and their probabilities averaged.
Baseline methods are trained on FF++ c23 following the standard protocol.
LOGER is trained on a multi-source pool (see Sec.~\ref{sec:setup}), so training data configurations differ from the baselines.

Table~\ref{tab:cross_effort} presents cross-dataset and cross-manipulation results following the evaluation of Effort~\cite{yan2024effort}.
The left group (Protocol-1) evaluates cross-dataset generalization on five benchmarks; the right group (Protocol-2) evaluates cross-manipulation generalization on eight forgery methods from DF40~\cite{yan2024df40} using their test splits.
Baseline results are directly cited from~\cite{yan2024effort}.
We note that multi-source training is integral to our framework design~\cite{wang2025scaling}, and that DFDC, DFDCP, and WDF are entirely absent from our training pool; for Protocol-2, all eight DF40 forgery methods are evaluated on test splits disjoint from our training data.
On Protocol-1, LOGER achieves an average AUC of 92.6\%, outperforming Effort (90.4\%) by +2.2\%.
The largest gains appear on DFDC (+6.0\%) and DFDCP (+3.4\%), both of which contain diverse manipulation types with significant compression artifacts, demonstrating the robustness and reliability of our approach under challenging conditions.
On Protocol-2, LOGER reaches an average of 96.4\%, surpassing Effort (94.0\%) by +2.4\% and ranking first on 6 out of 8 forgery methods.
Performance is especially strong on UniFace (99.1\%), e4s (98.9\%), and SimSwap (98.6\%), all of which are face-swapping methods that leave localized blending artifacts in specific facial regions.
These localized traces are particularly amenable to the local branch's patch-level MIL aggregation.

Table~\ref{tab:cross_gend} presents cross-dataset results on three benchmarks following the evaluation of GenD~\cite{yermakov2026deepfake}, with the backbone architecture and input type listed for each method.
Baseline results are directly cited from~\cite{yermakov2026deepfake}.
LOGER achieves the highest AUC on CDFv2 (95.9\%) and DFDC (90.3\%).
The DFDC result is noteworthy: LOGER outperforms ForAda (87.2\%), the previous best frame-level method, by +3.1 points on a dataset known for its diverse manipulation types, low-quality sources, and heavy compression.
On DFD, a near-saturated benchmark, LOGER achieves 96.2\% and remains competitive.

\begin{table}[t]
\centering
\caption{Comparison with representative methods on the NTIRE 2026 validation set (image-level AUC).}
\vspace{-2mm}
\small
\begin{tabular}{lc@{\hskip 12pt}lc}
\toprule
Method & AUC & Method & AUC \\
\midrule
GPT-5.2 & 0.43 & DDA~\cite{chen2025dualdataalignment} & 0.49 \\
Claude Sonnet 4.5 & 0.44 & MIRROR~\cite{liu2026mirror} & 0.59 \\
Gemini 3 Pro & 0.56 & Effort~\cite{yan2024effort} & 0.65 \\
CLIP-ViT-L~\cite{clip} & 0.66 & Reality Defender~\cite{realitydefender} & 0.70 \\
Qwen3-VL-2B~\cite{bai2025qwen3} & 0.73 & GenD~\cite{yermakov2026deepfake} & 0.74 \\
ConvNeXt-L~\cite{liu2022convnet} & 0.75 & \cellcolor{blue!5}\textbf{LOGER (Ours)} & \cellcolor{blue!5}\textbf{0.92} \\
\bottomrule
\end{tabular}
\label{tab:sota_compare}
\vspace{-2mm}
\end{table}

\paragraph{Competition validation set.}
Table~\ref{tab:sota_compare} compares LOGER against representative methods on the NTIRE 2026 competition validation set (image-level AUC).
Commercial MLLMs perform poorly regardless of adaptation strategy: zero-shot GPT-5.2 (43\%), Claude Sonnet 4.5 (44\%), and Gemini 3 Pro (56\%) all fall near or below chance, and even fine-tuned Qwen3-VL (73\%) remains below specialized approaches, confirming that general-purpose reasoning is insufficient under heavy degradation.
Among released detectors, DDA (49\%) and Effort (65\%) suffer from distribution shift, while GenD (74\%) benefits from diverse training.
CLIP-ViT-L (66\%) and ConvNeXt-L (75\%), trained on our 21K data, show that conventional backbones alone cannot close the gap.
LOGER achieves \textbf{92\%} AUC, outperforming the best baseline by +17 points.
On the official competition leaderboard, LOGER achieves 0.8901 AUC on the public test set and 0.8824 on the private test set, achieving 2nd place. The moderate drop from the validation set is expected, as both test splits introduce novel degradation types not seen during training; the consistent performance across the two test splits confirms strong generalization.

\subsection{Ablation Study}
\label{sec:ablation}

We conduct ablation experiments to analyze the contribution of each design dimension.

\paragraph{Per-model training strategies.}
Table~\ref{tab:ablation_permodel} presents progressive training enhancements for each backbone on the competition validation set.
For DINOv3-H (group a), starting from a baseline trained only on official data (0.5601 AUC), external data brings the largest single improvement (+0.211); Focal Loss adds +0.017; degradation augmentation contributes +0.058; and flip TTA provides +0.005.
M2, which increases inference resolution from 256 to 384 without retraining, adds another +0.007 over M1.
For MetaCLIP2-H (group b), M3 achieves 0.8008 AUC after incorporating external data, Focal Loss, and degradation augmentation.
Although its standalone accuracy is lower than DINOv3 models, MetaCLIP2's contrastive image-text pre-training encodes fundamentally different visual priors, contributing diversity rather than individual strength.
For DINOv3-L (group c), M4 achieves 0.8296 AUC with higher-resolution inference, and M5, fine-tuned from M4 at 338$\times$338, attains 0.8720 AUC, the highest among all individual models.

\begin{table}[t]
\centering
\small
\caption{Per-model ablation on the NTIRE 2026 competition validation set. Each group shows progressive training enhancements for one backbone.}
\vspace{-2mm}
\begin{tabular}{lc}
\toprule
Configuration & AUC \\
\midrule
\multicolumn{2}{l}{\textit{(a) DINOv3-H (Global)}} \\
Baseline (official data only) & 0.5601 \\
+ External data (ED) & 0.7714 \\
+ Focal Loss (FL) & 0.7882 \\
+ Degradation augmentation (DG) & 0.8464 \\
+ Flip TTA \textbf{(M1)} & 0.8515 \\
+ Higher-res inference 384 \textbf{(M2)} & \textbf{0.8584} \\
\midrule
\multicolumn{2}{l}{\textit{(b) MetaCLIP2-H (Global)}} \\
Baseline & 0.7190 \\
+ External data (ED) & 0.7562 \\
+ Focal Loss (FL) & 0.7767 \\
+ Degradation augmentation \textbf{(M3)} & \textbf{0.8008} \\
\midrule
\multicolumn{2}{l}{\textit{(c) DINOv3-L (Local)}} \\
Baseline & 0.7316 \\
+ External data (ED) & 0.8192 \\
+ Higher-res inference 384 \textbf{(M4)} & 0.8296 \\
+ Further fine-tuning 338 \textbf{(M5)} & \textbf{0.8720} \\
\bottomrule
\end{tabular}
\label{tab:ablation_permodel}
\vspace{-2mm}
\end{table}

\paragraph{Ensemble composition and fusion strategy.}
Table~\ref{tab:ablation_ensemble} analyzes the ensemble on the public test set from three perspectives: global branch effectiveness, local branch effectiveness, and fusion strategy.
The global-only sub-ensemble (M1+M2+M3) reaches 0.8812 AUC, showing that backbone heterogeneity yields clear gains over any single global model.
The local sub-ensemble (M4+M5) reaches 0.8612 AUC, confirming that patch-level MIL top-$k$ aggregation is a strong detection paradigm on its own.
The full five-model ensemble with logit averaging achieves \textbf{0.8901} AUC, substantially outperforming both the global-only and local-only sub-ensembles, directly validating the local--global complementarity that is central to our framework design.
Comparing fusion strategies, logit averaging (0.8901) outperforms probability averaging (0.8887) and majority voting (0.8812), confirming that fusing in the logit space before the sigmoid preserves inter-model disagreements and yields more robust predictions.

\begin{table}[t]
\centering
\caption{Ablation study on ensemble composition and fusion strategy on the NTIRE 2026 public test set. All results are image-level AUC.}
\label{tab:ablation_ensemble}
\small
\vspace{-2mm}
\begin{tabular}{lc}
\toprule
Configuration & AUC \\
\midrule
\multicolumn{2}{l}{\textit{(a) Global branch}} \\
M1 (DINOv3-H, 256) & 0.8674 \\
M2 (DINOv3-H, 384) & 0.8684 \\
M3 (MetaCLIP2-H, 378) & 0.8465 \\
M1+M2+M3 & 0.8812 \\
\midrule
\multicolumn{2}{l}{\textit{(b) Local branch}} \\
M4 (DINOv3-L, 384) & 0.8437 \\
M5 (DINOv3-L, 338) & 0.8487 \\
M4+M5 & 0.8612 \\
\midrule
\multicolumn{2}{l}{\textit{(c) Full ensemble (M1--M5)}} \\
Majority voting & 0.8812 \\
Probability averaging & 0.8887 \\
\rowcolor{blue!5}
Logit averaging (Ours) & \textbf{0.8901} \\
\bottomrule
\end{tabular}
\vspace{-2mm}
\end{table}

\subsection{Further Analysis}
\label{sec:further}

\paragraph{VFM backbone capability.}
We train a single DINOv3-Huge with cross-entropy loss on HydraFake~\cite{tan2025veritas} and evaluate on its hierarchical test protocol.
As shown in Table~\ref{tab:hydrafake}, even without Focal Loss, degradation augmentation, or ensemble, it achieves 99.4\% averaged accuracy, surpassing Veritas (90.7\%) and Co-SPY (84.7\%) by a large margin, justifying our choice of VFM backbones.

\begin{table}[t]
\centering
\caption{Accuracy (\%) on the HydraFake benchmark following the hierarchical evaluation protocol of Veritas~\cite{tan2025veritas}. All methods are trained on the HydraFake training set. ID: In-Domain, CM: Cross-Model, CF: Cross-Forgery, CD: Cross-Domain.}
\vspace{-1mm}
\small
\begin{tabular}{lccccc}
\toprule
Method & ID & CM & CF & CD & Avg. \\
\midrule
\multicolumn{6}{l}{\textit{Small vision models}} \\
D$^3$~\cite{yang2025d} & 87.3 & 93.6 & 73.7 & 73.1 & 81.1 \\
Effort~\cite{yan2024effort} & 94.7 & 82.8 & 88.0 & 67.1 & 82.2 \\
Co-SPY~\cite{cheng2025co} & 86.3 & 93.5 & 87.3 & 74.7 & 84.7 \\
\midrule
\multicolumn{6}{l}{\textit{MLLM-based detectors}} \\
SIDA-7B~\cite{huang2025sida} & -- & 76.0 & 63.3 & 69.9 & 76.3 \\
FakeVLM~\cite{wen2025spot} & -- & 77.0 & 77.7 & 75.7 & 77.3 \\
Veritas~\cite{tan2025veritas} & 97.3 & 94.8 & 91.7 & 82.2 & 90.7 \\
\midrule
\rowcolor{blue!5}
\textbf{DINOv3-H} & \textbf{99.5} & \textbf{99.9} & \textbf{99.8} & \textbf{98.4} & \textbf{99.4} \\
\bottomrule
\end{tabular}
\label{tab:hydrafake}
\vspace{-1mm}
\end{table}

\begin{figure*}[t]
    \centering
    \includegraphics[width=\linewidth]{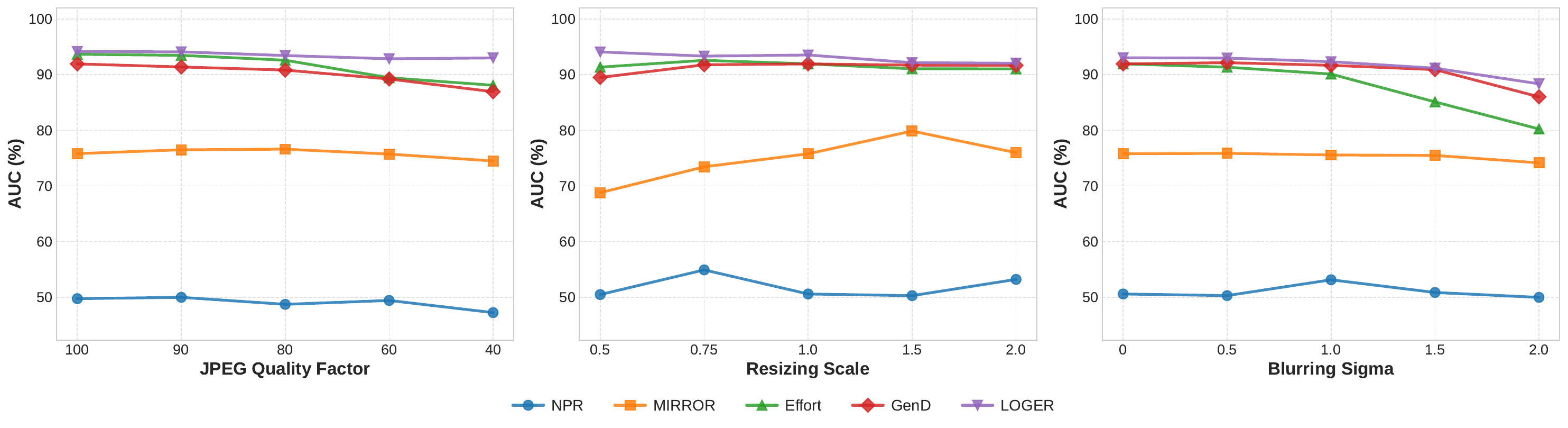}
    \caption{Robustness analysis under three degradation types: JPEG compression (left), spatial resizing (middle), and Gaussian blurring (right). AUC (\%) is computed over 1,000 videos sampled from five cross-dataset benchmarks.}
    \label{fig:robust}
    \vspace{-1mm}
\end{figure*}

\paragraph{Robustness under degradation.}
We construct a controlled robustness benchmark of 1,000 videos (100 real + 100 fake from each of the five test sets) and apply JPEG compression (QF 100$\to$40), spatial resizing (scale 0.5$\to$2.0), and Gaussian blurring ($\sigma$ 0$\to$2.0) independently at varying severity levels.

Fig.~\ref{fig:robust} compares LOGER with four baselines spanning different detection paradigms.
NPR~\cite{tan2024rethinking}, based on neighboring pixel relationships, hovers near chance (${\sim}50\%$) across all conditions.
MIRROR~\cite{liu2026mirror} stays in the $69$--$80\%$ range, showing limited effectiveness.
Effort and GenD both start above $91\%$ under clean conditions but degrade noticeably under severe perturbations: GenD drops from $91.9\%$ to $86.9\%$ under heavy JPEG compression (QF=40), and Effort falls from $92.0\%$ to $80.2\%$ under strong Gaussian blurring ($\sigma$=2.0), an $11.7$ point decline.
LOGER consistently achieves the highest AUC across all degradation types and severity levels.
Under JPEG compression, it drops only $1.1$ points; under resizing, it remains above $92.0\%$ at all scales; under blurring, its largest drop is $4.7$ points (from $93.0\%$ to $88.4\%$ at $\sigma$=2.0), less than half of Effort's decline under the same condition.
These results demonstrate the strong robustness of LOGER under diverse real-world degradation conditions.

\begin{figure}[t]
    \centering
    \includegraphics[width=\linewidth]{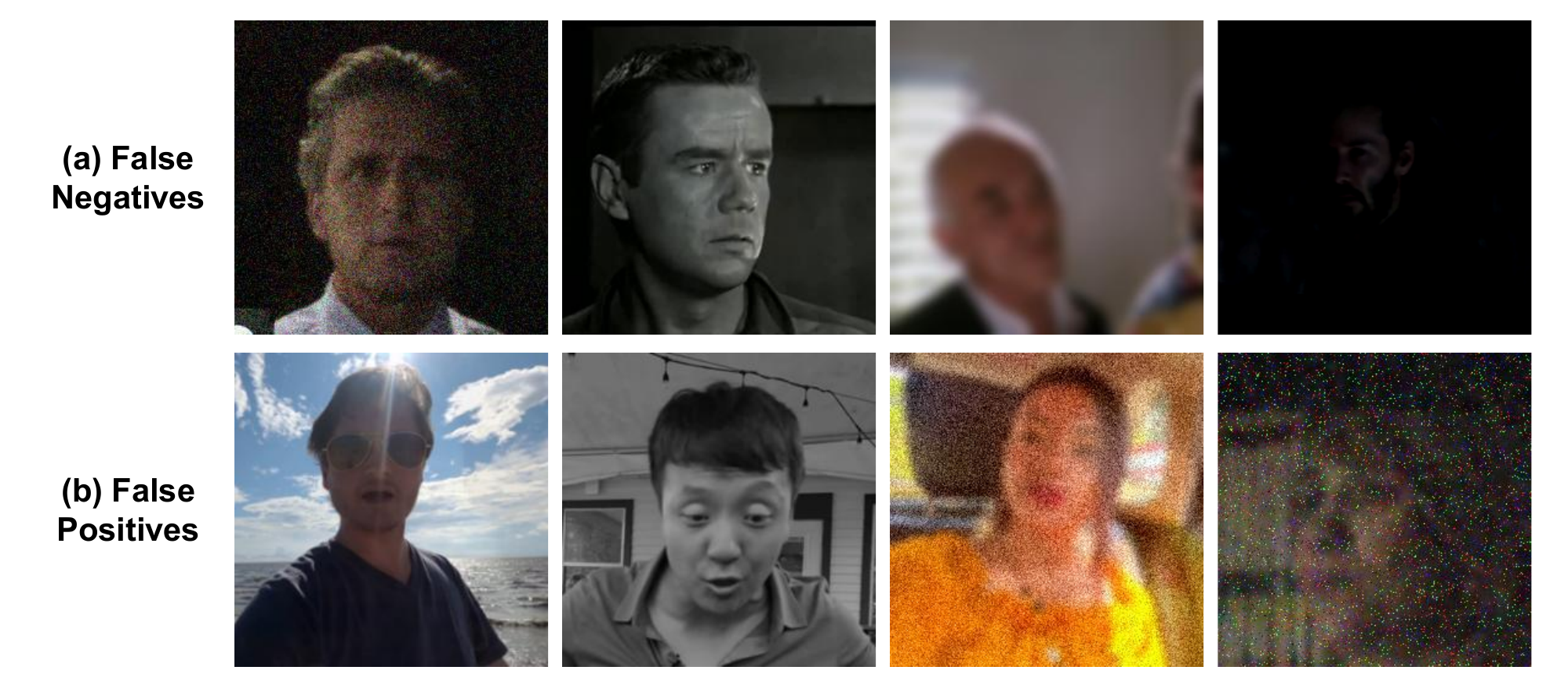}
    \caption{Representative failure cases on the NTIRE 2026 public test set. Top: false negatives (fake$\to$real). Bottom: false positives (real$\to$fake).}
    \label{fig:badcase}
    \vspace{-1mm}
\end{figure}

\paragraph{Failure analysis.}
Fig.~\ref{fig:badcase} shows representative failure cases from the NTIRE 2026 public test set.
False negatives (top) are fake images misclassified as real, typically suffering from heavy noise, severe blur, extreme darkness, or grayscale conversion that destroys the forgery traces the detector relies on.
False positives (bottom) are real images misclassified as fake, where strong lens flare, color noise, or heavy compression introduces visual patterns that mimic generation artifacts.
Both failure modes share the same root cause: extreme degradation erases the boundary between real and fake distributions, making reliable discrimination difficult.

\section{Conclusion}
\label{sec:conclusion}

We present LOGER, a local--global ensemble framework for robust deepfake detection.
The global branch captures holistic anomalies via heterogeneous VFM backbones at multiple resolutions, while the local branch mitigates evidence dilution through MIL top-$k$ aggregation with dual-level supervision.
Logit-space fusion exploits the decorrelated errors of the two branches for stable predictions.
Experiments across multiple cross-dataset benchmarks, the NTIRE 2026 Robust Deepfake Detection Challenge, and controlled robustness stress tests consistently demonstrate strong generalization and robustness under diverse manipulation methods and real-world degradations.
Our failure analysis identifies extreme degradation as the primary remaining bottleneck, where discriminative information between real and fake content is largely destroyed.
Future work will explore degradation-adaptive inference strategies that explicitly account for input quality during detection.

\clearpage
{
    \small
    \bibliographystyle{ieeenat_fullname}
    \bibliography{main}
}


\end{document}